\documentclass[journal]{IEEEtran}
\usepackage{graphicx}
\usepackage{array,booktabs}
\usepackage{multirow}
\usepackage{tabularx}
\usepackage{bm}
\usepackage{amsmath}
\usepackage{amsfonts}
\usepackage{amssymb}
\usepackage{subfigure}
\usepackage{epstopdf}
\usepackage{xcolor,colortbl}
\usepackage{cite}
\usepackage{dblfloatfix}
\usepackage{tikz}
\usetikzlibrary{shapes,arrows,fit,positioning}

\newcommand{\cellc}{\cellcolor{gray!20}}
\DeclareMathOperator*{\argmax}{argmax}
\newcommand\RotText[1]{\rotatebox{90}{\parbox{2.5cm}{\centering#1}}}
\newcolumntype{C}[1]{>{\centering\let\newline\\\arraybackslash\hspace{0pt}}m{#1}}

\begin{document}

\title{Information Theoretic Feature Transformation Learning for Brain Interfaces}

\author{Ozan~\"{O}zdenizci,~\IEEEmembership{Student~Member,~IEEE,} and~Deniz~Erdo\u{g}mu\c{s},~\IEEEmembership{Senior~Member,~IEEE}%
\thanks{O.~\"{O}zdenizci and D.~Erdo\u{g}mu\c{s} are with the Cognitive Systems Laboratory at Department of Electrical and Computer Engineering, Northeastern University, Boston, MA, USA. E-mail: \{oozdenizci, erdogmus\}@ece.neu.edu.}%
\thanks{Our work is supported by NSF (IIS-1149570, CNS-1544895, IIS-1715858), DHHS (90RE5017-02-01), and NIH (R01DC009834).}%
\thanks{Copyright (c) 2017 IEEE. Personal use of this material is permitted. However, permission to use this material for any other purposes must be obtained from the IEEE by sending an email to pubs-permissions@ieee.org.}%
}

\markboth{IEEE Transactions on Biomedical Engineering}%
{\"{O}zdenizci \MakeLowercase{\textit{et al.}}: Information Theoretic Feature Transformation Learning for Brain Interfaces}

\maketitle


\begin{abstract}\textit{Objective:} A variety of pattern analysis techniques for model training in brain interfaces exploit neural feature dimensionality reduction based on feature ranking and selection heuristics. In the light of broad evidence demonstrating the potential sub-optimality of ranking based feature selection by any criterion, we propose to extend this focus with an information theoretic learning driven feature transformation concept. \textit{Methods:} We present a maximum mutual information linear transformation (MMI-LinT), and a nonlinear transformation (MMI-NonLinT) framework derived by a general definition of the feature transformation learning problem. Empirical assessments are performed based on electroencephalographic (EEG) data recorded during a four class motor imagery brain-computer interface (BCI) task. Exploiting state-of-the-art methods for initial feature vector construction, we compare the proposed approaches with conventional feature selection based dimensionality reduction techniques which are widely used in brain interfaces. Furthermore, for the multi-class problem, we present and exploit a hierarchical graphical model based BCI decoding system. \textit{Results:} Both binary and multi-class decoding analyses demonstrate significantly better performances with the proposed methods. \textit{Conclusion:} Information theoretic feature transformations are capable of tackling potential confounders of conventional approaches in various settings. \textit{Significance:} We argue that this concept provides significant insights to extend the focus on feature selection heuristics to a broader definition of feature transformation learning in brain interfaces.\end{abstract}


\IEEEpeerreviewmaketitle

\section{Introduction}

\IEEEPARstart{O}{ver} the last decades, electroencephalogram (EEG) based brain-computer interfaces (BCIs) have shown the promise of providing a direct neural communication and control channel in paralysis, and reinforce motor restoration in stroke \cite{Wolpaw:2002,Birbaumer:2007}. In the design of closed-loop brain/neural interfaces for people with neuromuscular disabilities, a variety of statistical signal processing and pattern analysis approaches have been considered. For supervised neural decoding model construction, improving generalization and optimal exploitation of the information content in the extracted neural features with respect to their class conditions (i.e., labels) is essential given a finite number of training data samples. Furthermore, the number of daily training examples will be strictly limited for patients with severe neuromuscular disabilities, due to the constrained data collection times under adequate concentration and consciousness. This optimal exploitation of extracted features can be performed with various feature learning and dimensionality reduction frameworks, which enables elimination or weighting of redundant features that do not convey reliable statistical information for decoding, or avoid overfitting by reducing the constructed model complexity.

A theoretically optimal dimensionality reduction procedure given a set of training examples and a specified classifier will point to iteratively adjusting a pre-determined feature learning framework until the best cross validated classification accuracy is achieved, which are known as the \textit{wrapper} approaches. Since this is naturally unfeasible, \textit{filter} approaches provide an alternative for optimizing a feature learning framework based on an optimality criterion. Specifically, feature ranking and subset selection algorithms \cite{Guyon:2003}, and particularly feature selection based on information theoretic criteria, where salient statistical properties of features can be exploited in the form a probabilistic dependence measure, have shown significant promise \cite{Battiti:1994,Kwak:2002}. Likewise, a vast body of contemporary brain interfaces rely on subject-specific feature selection methods for feature dimensionality reduction \cite{Garrett:2003,Muller:2004,Lal:2004,Krusienski:2008,Tomioka:2010}, particularly based on maximum mutual information criterion \cite{Ang:2012,Muhl:2014,Jenke:2014}, which were also investigated by two extensive, recent and complementary survey studies on BCIs \cite{Lotte:2007,Lotte:2018}.

In other respects, there exists significant evidence claiming that feature ranking by any criterion, including information theoretic criteria, being potentially sub-optimal \cite{Erdogmus:2008,Torkkola:2008}. This argument can simply be extended from statistical demonstrations on any two redundant features being informative jointly, or that high correlation between features should not necessarily be interpreted as lack of feature complementarity \cite{Guyon:2003}. Based on this idea, information theoretic feature projection approaches are introduced in the form of linear projections \cite{Torkkola:2003,Chen:2008,Zhang:2010,Faivishevsky:2012,Nenadic:2007} or rotation matrices \cite{Hild:2006}. This feature transformation approach can be interpreted as determining a manifold on which projections of the original extracted features carry maximal mutual information with the class labels. However, these specific approaches are in need of computationally feasible practical approximations, and they are not yet considered for feature learning in brain interfaces. We argue that exploiting such an approach may provide significant insights, particularly for multi-class BCIs which are more inclined to overfitting with high-dimensional training feature spaces, and sub-optimal feature selection based dimensionality reduction confounders. From a neurophysiological standpoint, feature projection approaches align with the widely-acknowledged hypothesis that distributed networks of cortical sources are likely to generate brain responses that are associated with specific tasks \cite{Mantini:2007,Bressler:2010}. Hence, BCI decoder models could potentially excel from arbitrary synergies of extracted EEG features representative of various neural activities, rather than a selected subset.

In this article, we propose a general definition for information theoretic feature transformation learning, which we stochastically estimate on finite training data sets for feature extraction in brain interfaces. We present a maximum mutual information linear transformation (MMI-LinT) approach, which we previously evaluated in binary decoding \cite{Ozdenizci:2017b}, and a nonlinear transformation (MMI-NonLinT) approach derived by the general definition. Furthermore, we introduce a graphical model based hierarchical multi-class decoding framework, which can be considered as an intuitively specified case of one-versus-rest binary classifiers. We argue that a hierarchical binary feature transformation learning approach in this multi-class framework is likely to outperform heuristic feature selection algorithms. We empirically assess MMI-LinT and MMI-NonLinT using EEG data recorded during a cue-based four class motor imagery BCI task \cite{Pfurtscheller:2001}. Firstly, we exploit state-of-the-art methods for initial feature vector construction; common spatial patterns (CSP) \cite{Ramoser:2000,Blankertz:2008} and filter bank CSP (FBCSP) extensions \cite{Ang:2008}. Subsequently, we compare our feature learning and dimensionality reduction approach with both statistical testing based, as well as mutual information based feature ranking and selection methods explored in previous BCI studies. Finally, we discuss the significance of our results and provide insights that extends the feature selection based focus to feature transformation learning in brain interfaces.

\section{Information Theoretic Feature Transformation Learning}
\label{sec:itfl}

In this section, we introduce the information theoretic feature transformation learning objective. We discuss mutual information in Bayesian optimal classification, present the stochastic estimation approach for the objective, and introduce the linear and nonlinear transformation schemes.

\subsection{Objective Formulation}
\label{sec:formulation}

Let $\{\bm{x}_i\}_{i=1}^{n}\subseteq\mathbb{R}^{d_x}$ denote the observational finite data set consisting of $n$ samples of a continuous valued random variable $\mathit{X}$, where $\bm{x}_i$ is the $d_x$-dimensional feature vector (e.g., pre-processed EEG data) representing the $i$-th sample. Likewise, let $\{c_i\}_{i=1}^{n}$ denote the set of their respective class labels consisting of $n$ samples of a discrete valued random variable $\mathit{C}$, where each $c_i$ represents the class category varying between $1$ to $L$, with $L$ being the number of classes. The objective in the learning problem is to find a transformation $\psi^\star:\mathbb{R}^{d_x}\mapsto\mathbb{R}^{d_y}$ that maps the $d_x$-dimensional input feature space to a $d_y$-dimensional transformed feature space, while maximizing the mutual information between the transformed data and corresponding class labels, based on the observational finite data set samples:
\begin{equation}
    \psi^\star = \argmax_{\psi\in\Omega} \{I(\mathit{Y},\mathit{C})\},
    \label{eq:objective}
\end{equation}
with continuous random variable $\mathit{Y}$ having transformed data set samples $\bm{y}_i=\psi^\star(\bm{x}_i;\bm{\theta}^\star)$ in the $d_y$-dimensional feature space, $\bm{\theta}$ denoting the parameters of the function $\psi$, $I(\mathit{Y},\mathit{C})$ the mutual information between random variables $\mathit{Y}$ and $\mathit{C}$, and $\Omega$ the feature transform function space. We will denote the probability density for the random variable $\mathit{Y}$ with $p(\bm{y})$, and probability mass function for $\mathit{C}$ with $P(c)$. We will assume $d_x<d_y$ for dimensionality reduction in model training.

\subsection{Information Theoretic Bounds on Classification Error}
\label{sec:bayesbounds}

In Bayesian optimal classification, upper and lower bounds on the probability of error $P_e$ in estimating a discrete valued random variable $\mathit{C}$ from an observational random variable $\mathit{Y}$ can be derived by information theoretic criteria. Using the notation we provided above, for a binary classification problem, these bounds can be determined as:
\begin{equation}
    \frac{H(\mathit{C})-I(\mathit{Y},\mathit{C})}{2}\ge P_e\ge\frac{H(\mathit{C})-I(\mathit{Y},\mathit{C})-1}{\text{log}(2)},
    \label{eq:bounds}
\end{equation}
with $P_e=P(c\ne\widehat{c})$ where $\hat{c}$ is the predicted class label while estimating $c$ after observing a sample of $\mathit{Y}$, and $H(.)$ is Shannon's entropy. In Eq.~\ref{eq:bounds}, lower bound to the probability of error is given by Fano's inequality \cite{Fano:1961}, and the upper bound on Bayes error is known as the Hellman-Raviv bound \cite{Hellman:1970}. Together, these inequalities claim that the lowest possible Bayes error of any given classifier providing the class label prediction $\hat{c}$ can be achieved when the mutual information between the random variables $\mathit{Y}$ and $\mathit{C}$ is maximized.

\subsection{Stochastic Mutual Information Gradient}
\label{sec:smig}

Mutual information between the continuous random variable $\mathit{Y}$ and the discrete class labels random variable $\mathit{C}$ is defined as: $I(\mathit{Y},\mathit{C}) = H(\mathit{Y}) - H(\mathit{Y}\vert\mathit{C})$. It is important to note that estimating Eq.~\ref{eq:objective} is a challenging problem, as recently studied \cite{Ross:2014,Gao:2017}, since it includes both continuous and discrete random variables where the entropy of a continuous random variable can have infinitely large positive or negative values, whereas the entropy of a discrete random variable is always positive. Formally, the mutual information is denoted as:
\begin{equation}
    \begin{split}
    I(\mathit{Y},\mathit{C}) = & - \int_{\bm{y}} p(\bm{y})\log p(\bm{y})d\bm{y} \\ & + \int_{\bm{y}} \sum_{c} p(\bm{y},c)\log p(\bm{y} \vert c)d\bm{y}.
    \end{split}
    \label{eq:mutualinfo}
\end{equation}

We will approach the optimization problem stochastically based on the observational data set samples and their corresponding class labels. In this context, precise estimation of mutual information is not necessary, but we aim to adaptively estimate the optimal feature transformation function parameters under maximum mutual information criterion. This approach is motivated by similar work on stochastic entropy and mutual information estimation models \cite{Erdogmus:2003,Chen:2008}.

In our adaptive algorithm, parameters $\bm{\theta}$ will be iteratively updated based on the instantaneous estimate of the gradient of mutual information at each iteration $t$ (i.e., $\nabla_{\bm{\theta}}\widehat{I}_t(\mathit{Y},\mathit{C})=\partial\widehat{I}_t(\mathit{Y},\mathit{C})/\partial\bm{\theta}$), which we will refer to as the \textit{stochastic mutual information gradient}. Here, in fact, we approximate the true gradient of the mutual information (i.e., $\nabla_{\bm{\theta}}I(Y,C)$) stochastically, and perform gradient ascent parameter updates based on the instantaneous gradient estimate $\nabla_{\bm{\theta}}\widehat{I}_t(\mathit{Y},\mathit{C})$ evaluated with the instantaneous sample $\bm{y}_t$ and the value of $\bm{\theta}$ at iteration $t$. This stochastic quantity can be obtained by dropping the expectation operation over $\mathit{Y}$ from:
\begin{equation}
    \begin{split}
    \nabla_{\bm{\theta}}I(\mathit{Y},\mathit{C}) = & \frac{\partial}{\partial\bm{\theta}} \Bigg[ - \int_{\bm{y}} p(\bm{y})\log p(\bm{y})d\bm{y} \\ & + \int_{\bm{y}} p(\bm{y}) \sum_{c} P(c \vert \bm{y})\log p(\bm{y} \vert c)d\bm{y} \Bigg],
    \end{split}
    \label{eq:mutualinfograd}
\end{equation}
such that the resulting expression (i.e., stochastic mutual information gradient at iteration $t$) will be expressed by:
\begin{equation}
    \begin{split}
    \nabla_{\bm{\theta}}\widehat{I}_t(\mathit{Y},\mathit{C}) = & \frac{\partial}{\partial\bm{\theta}} \Bigg[- \log \widehat{p}(\bm{y}_t) \\ & +  \sum_{c} \widehat{P}(c \vert \bm{y}_t) \log \widehat{p}(\bm{y}_t \vert c)\Bigg].
    \end{split}
    \label{eq:instmutualinfograd}
\end{equation}

In practice, the probability density for $\mathit{Y}$ is not known, hence we can non-parametrically approximate by kernel density estimations in the form of $\widehat{p}(\bm{y}) = (1/n) \sum_{i=1}^{n} \bm{\kappa}_\sigma(\bm{y}-\bm{y}_i)$ with $\bm{\kappa}_\sigma(.)$ being the size $\sigma$ multivariate kernel function for a $d_y$ dimensional vector \cite{Principe:2000}. Note that a continuously differentiable kernel is necessary for proper evaluation of the gradients. Here, the stochastic estimator in Eq.~\ref{eq:instmutualinfograd} is a biased estimator of the actual mutual information gradient in Eq.~\ref{eq:mutualinfograd}, since it is based on kernel density estimators with finite samples which are biased estimators \cite{Parzen:1962}. Going further, applying the Bayes' Theorem, the stochastic mutual information estimate $\widehat{I}_t(\mathit{Y},\mathit{C})$ from Eq.~\ref{eq:instmutualinfograd} can be expressed by:
\begin{equation}
    \begin{split}
    \widehat{I}_t(\mathit{Y},\mathit{C}) = & -\log \left( \sum_{c} \widehat{P}(c)\widehat{p}(\bm{y}_t \vert c) \right) \\ & + \sum_{c} \left(\frac{\widehat{P}(c)\widehat{p}(\bm{y}_t \vert c)}{\sum_{c} \widehat{P}(c)\widehat{p}(\bm{y}_t \vert c)} \right) \log \widehat{p}(\bm{y}_t \vert c),
    \end{split}
    \label{eq:smig}
\end{equation}
where $\widehat{p}(\bm{y}_t \vert c)$ at each iteration $t$ can be estimated either parametrically (e.g., Gaussian) or non-parametrically through Gaussian kernel density fitting on class conditional distributions of the transformed training data, and class priors $\widehat{P}(c)$ can be determined again on the training data samples. 

During model training, we employ momentum stochastic gradient ascent \cite{Qian:1999}. Parameter update $\bm{u}_t$ at iteration $t$ is determined by $\bm{u}_t = \gamma\bm{u}_{t-1}+\eta\nabla_{\bm{\theta}}\widehat{I}_t(\mathit{Y},\mathit{C})$, which further is employed as $\bm{\theta}\leftarrow\bm{\theta} + \bm{u}_t$, where $\gamma$ is the momentum parameter and $\eta$ is the step size for the gradient. Iterations are performed using all training data samples in randomized order with a batch size of one sample, and also repeated for a number of training epochs. Choice of the function $\psi$ and its parameters specifies the feature transformation scheme. In this paper, we propose a linear (MMI-LinT) and a nonlinear (MMI-NonLinT) transformation modality as presented henceforth.

\subsection{Linear Feature Transformation (MMI-LinT)}
\label{sec:mmilint}

In the MMI-LinT framework, transformation function is parameterized by a linear projection matrix. At each iteration $t$, the transformation function $\psi$ generates samples of the random variable $\mathit{Y}$ according to $\psi(\bm{x}_t;\bm{\mathit{M}})=\bm{\mathit{M}}\bm{x}_t=\bm{y}_t$, where elements of the matrix $\bm{\mathit{M}}$ of size $d_y\times d_x$ are updated by the adaptive linear system. Accordingly, the stochastic mutual information gradient can be denoted as:
\begin{equation}
    \nabla_{\bm{\mathit{M}}}\widehat{I}_t(Y,C) = 
    \left(\nabla_{\bm{y}_t}\widehat{I}_t(Y,C)\right)\cdot\bm{x}_t,
    \label{eq:mmilint}
\end{equation}
with $\bm{x}_t$ one of the data set samples from $\{\bm{x}_i\}_{i=1}^{n}$ during model training. Using continuously differentiable class conditional kernel density estimations or a parametric density, the gradients with respect to $\bm{y}_t$ can be obtained numerically. From a computational implementation perspective, this simply corresponds to backpropagation of $\widehat{I}_t(\mathit{Y},\mathit{C})$ through a single fully-connected layer neural network. Dimensionality of linear projection $d_y$ remains as a parameter to be determined, alongside the number of training epochs.

\subsection{Nonlinear Feature Transformation (MMI-NonLinT)}
\label{sec:mminonlint}

A nonlinear transformation function can be parameterized in various modalities. We employ a muiltilayer perceptron framework for MMI-NonLinT. Specifically in a two-layer perceptron case, which we employed in our demonstrations (c.f.~Section~\ref{sec:results}), the transformation function is denoted as a combination of a linear input projection with a nonlinear activation function that outputs to the hidden layer, and a linear output layer projection. For the hidden layer nonlinearity we use a rectified linear unit (ReLU) transfer function.

Overall, by definition of the presented two-layer perceptron network, nonlinear feature transformation at iteration $t$ can be formulated as a composition of the transformation functions $g_1:\mathbb{R}^{d_x}\mapsto\mathbb{R}^{d_z}$ where $d_z$ denotes the hidden layer dimensionality, and $g_2:\mathbb{R}^{d_z}\mapsto\mathbb{R}^{d_y}$. These can be represented as:
\begin{equation}
    \begin{split}
    g_1(\bm{x}_t) &= \bm{z}_t = \text{max}(0,\bm{\mathit{M}}_1\bm{x}_t), \\
    g_2(\bm{z}_t) &= \bm{y}_t = \bm{\mathit{M}}_2\bm{z}_t, \\
    \psi(\bm{x}_t) &= g_2\left(g_1(\bm{x}_t)\right)=\bm{y}_t.
    \end{split}
    \label{eq:mminonlint}
\end{equation}

During MMI-NonLinT feature learning implementations, stochastic mutual information gradients can be estimated by backpropagating $\widehat{I}_t(\mathit{Y},\mathit{C})$ through the multilayer perceptron, to iteratively estimate the optimal projection matrices $\bm{\mathit{M}}_1^\star$ and $\bm{\mathit{M}}_2^\star$. Number of nodes in the hidden layer (i.e., dimensionality $d_z$ of projection $\bm{\mathit{M}}_1$) is another parameter to be determined alongside $d_y$ and number of training epochs.

\section{Hierarchical Multi-Class Decoding}
\label{sec:hierarchicaldecoding}

In this section, we introduce the binary hierarchical classification scheme we employ for multi-class decoding. We present a graphical model based representation, express the Bayesian decision criterion, and provide a coherent extension for the proposed information theoretic feature learning protocol.

\subsection{Hierarchical Graphical Model}
\label{sec:graphmodel}

We decompose the multi-class ($L$ class) problem into $L-1$ binary sub-problems. This results in a hierarchically arranged tree with $L-1$ one-versus-rest classifiers. In the context of BCIs, we argue that hierarchical arrangement of one-versus-rest binary sub-problems can be represented by an intuitive ordering rather than an arbitrary one. For instance in hand gesture decoding, upper hierarchical levels can discriminate the choice of hand and palm opening, whereas lower levels will be decoding power versus precision grasp type, or thumb abduction versus adduction of a specific grasp \cite{Ozdenizci:2018}. This decomposition provides an application specific multi-class decoding scheme which we demonstrate in Section~\ref{sec:results}.

\begin{figure}
    \centering
    \begin{tikzpicture}[thick,scale=0.47, every node/.style={transform shape}]
    
    \tikzstyle{response} = [cloud, circle, draw, fill=blue!10, 
        text width=6em, text centered,  minimum height=3em]
    \tikzstyle{level1} = [cloud, circle, draw, fill=white!10, 
        text width=6em, text centered,  minimum height=2em] 
    \tikzstyle{desire} = [cloud, circle, draw, fill=red!10, 
        text width=6em, text centered,  minimum height=2em] 
    \tikzstyle{dots} = [cloud, circle, text width=6em, text centered, minimum height=2em] 
    \tikzstyle{arrow} = [thick,->,>=stealth]
    
    \node [level1] (S1) {\Huge $S_i^{(1)}$};
    \node [level1, right = 1.4cm of S1] (S2) {\Huge $S_i^{(2)}$};
    \node [level1, right = 1.4cm of S2] (S3) {\Huge $S_i^{(3)}$};
    \node [level1, right = 3.5cm of S3] (SM) {\Huge $S_i^{(L-1)}$};
    \node [desire, above = 1.6cm of S2] (Decision) {\Huge $\widehat{c}_i$};
     
    \node [response, below = 1.2cm of S1] (EEGS1) {\Huge $\bm{y}_{i}^{(1)}$};
    \node [response, below = 1.2cm of  S2] (EEGS2) {\Huge $\bm{y}_{i}^{(2)}$};
    \node [response, below = 1.2cm of  S3] (EEGS3) {\Huge $\bm{y}_{i}^{(3)}$};
    \node [response, below = 1.2cm of  SM] (EEGSM) {\Huge $\bm{y}_{i}^{(L-1)}$};
    \node [dots, right = 0.65cm of S3] (DOTS) {\Huge$\bm{\ldots}$};

    \draw [arrow,dashed] (Decision) -- (S1);
    \draw [arrow,dashed] (Decision) -- (S2);
    \draw [arrow,dashed] (Decision) -- (S3);
    \draw [arrow,dashed] (Decision) -- (SM);
    \draw [arrow] (S1) -- (S2);
    \draw [arrow] (S2) -- (S3);
    \draw [arrow] (S3) -- (DOTS);
    \draw [arrow] (DOTS) -- (SM);
    
    \draw [arrow] (S1) -- (EEGS1);
    \draw [arrow] (S2) -- (EEGS2);
    \draw [arrow] (S3) -- (EEGS3);
    \draw [arrow] (SM) -- (EEGSM);

    \end{tikzpicture}
    \caption{Graphical representation of the decoding model. Features $\bm{y}_i^{(l)}$, extracted from observational data samples at level $l$, are indicated with blue nodes representing the observed random variables. Decision $\widehat{c}_i$ at sample $i$ is deterministically related (dashed lines) with the states at each level. Observed random variables are probabilistically related with the states (solid lines).}
    \vspace{-0.15cm}
    \label{graphicalmodel}
\end{figure}
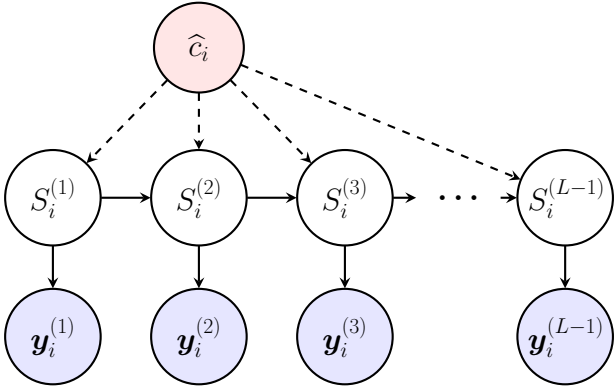

Hierarchical tree representation is depicted by the graphical model in Figure~\ref{graphicalmodel}. At each sample $i$, overall decision $\widehat{c}_i$ is deterministically related with the states at each level. Each state variable $S_i^{(l)}$ at level $l$ represents the decision for a binary sub-problem (i.e., $l=1$ to $l=L-1$ representing decisions from the highest to lowest levels). Extracted features $\bm{y}_i^{(l)}$ at level $l$ from observational data samples, are probabilistically related with sub-problem decisions. This hierarchical decoding approach can be interpreted as a special case of one-versus-rest multi-class decoding schemes with an intuitive ordering.

\subsection{Bayesian Decision Criterion}
\label{sec:bayesdecision}

Classification based on $\bm{y}_i$ (i.e., extracted features from observational data samples $\bm{x}_i$) is performed by maximum-a-posteriori (MAP) estimation. Relying on the graphical model and the hierarchical decomposition for level-wise feature extraction, MAP decision rule can be denoted as:
\begin{equation}
    \widehat{c}_i=\argmax_{c_i} P(c_i\vert\bm{y}_{i}^{(1)},\bm{y}_{i}^{(2)},\ldots,\bm{y}_{i}^{(L-1)}),
    \label{eq:decisioncriteria}
    \vspace{-0.2cm}
\end{equation}
with $\bm{y}_i^{(l)}$ denoting the extracted feature vector from a subset of observations only corresponding to level $l\in\{1,\ldots,L-1\}$. This ensures feature extraction to be performed between two classes at each level. Hence at the feature extraction step, the set $\{\bm{x}_i\}_{i=1}^{n}$ is split into one-versus-rest subsets based on the intuitive hierarchical ordering (c.f.~Section~\ref{sec:hitfl}). Based on the graphical model, Eq.~\ref{eq:decisioncriteria} can be denoted as:
\begin{equation}
    \widehat{c}_i=\argmax_{c_i} P(S_i^{(1)},\ldots,S_i^{(L-1)}\vert\bm{y}_{i}^{(1)},\ldots,\bm{y}_{i}^{(L-1)}),
    \label{eq:decisioncriteria2}
    \vspace{-0.1cm}
\end{equation}
which can further be represented by the across-level independency assumptions imposed by the graphical model as:
\begin{equation}
    \widehat{c}_i = \argmax_{c_i} \left\{\prod_{l=1}^{L-1} p(\bm{y}_i^{(l)} \vert S_i^{(l)}) P(S_i^{(l)} \vert S_i^{(l-1)}) \right\},
    \label{eq:decisioncriteria3}
    \vspace{-0.1cm}
\end{equation}
with the first expression in the product denoting the class conditional density of the extracted features at level $l$, second expression in the product denoting the class priors at level $l$, and $P(S_i^{(1)} \vert S_i^{(0)})=P(S_i^{(1)})$ for consistency. Eq.~\ref{eq:decisioncriteria3} can be evaluated for a test sample using the likelihoods based on class conditional kernel density estimations obtained with the training data for both $+1$ and $-1$ classes at all levels.

\subsection{Hierarchical Feature Transformation Learning}
\label{sec:hitfl}

We denote a consistent notation for combining the feature transformation learning approach with the hierarchical framework. For hierarchical feature extraction, $\bm{y}_i^{(1)}$ are obtained using the complete set $\{\bm{x}_i^{(1)}\}_{i=1}^{n}$, with corresponding binary labels $\{c_i^{(1)}\}_{i=1}^{n}$ based on the first level hierarchical disjunction. However, $\bm{y}_i^{(2)}$ are extracted based on the set $\{\bm{x}_i^{(2)}\}_{i=1}^{n_2}$ with labels $\{c_i^{(2)}\}_{i=1}^{n_2}$, where $n_2$ denotes the number of data samples that correspond to the second level hierarchical disjunction. In mathematical terms; $n=n_1=n_1^{+1} + n_1^{-1}$, $n_2 = n_1^{-1}$, $n_3 = n_2^{-1}$, and so on, with $n_l^{-1}$ denoting the number of samples with negative labels at level $l$ that consists all samples of level $l+1$. Here, the choice of $-1$ for the continuing disjunction branches was arbitrary.

For the information theoretic objectives, transformation functions are obtained at every hierarchical level, based on the subset of the data samples and their corresponding binary labels at that level. Overall, this can be denoted as:
\begin{equation}
    \psi^{(l)\star} = \argmax_{\psi^{(l)}\in\Omega} \{I(\mathit{Y}^{(l)},\mathit{C}^{(l)} \vert \mathit{C}^{(l-1)},\ldots,\mathit{C}^{(1)})\},
    \label{eq:hierobjective}
    \vspace{-0.2cm}
\end{equation}
with $\mathit{Y}^{(l)}$ a continuous valued random variable having transformed data set samples $\bm{y}_i^{(l)}=\psi^{(l)\star}(\bm{x}_i^{(l)};\bm{\theta}^{(l)\star})$ for level $l$, $\bm{\theta}^{(l)\star}$ denoting the parameters of the transformation function at level $l$, and $\mathit{C}^{(l)}$ a binary random variable for level $l$.

\section{Experimental Results}
\label{sec:results}

In this section, we implement and demonstrate feasibility of our approach using EEG data recorded during a cue-based four class motor imagery BCI task \cite{Pfurtscheller:2001}. For empirical assessments, we used data set 2a of the BCI Competition IV\footnote{BCI Competition IV: \url{http://www.bbci.de/competition/iv/}}\cite{Tangermann:2012}, which was provided by the Institute of Neural Engineering, Technische Universit\"{a}t Graz, Austria. We compare and discuss the results with conventional feature dimensionality reduction methods accordingly in binary and multi-class decoding.

\subsection{Study Design}

Nine healthy subjects (4 female; mean age~=~23.11$\pm$2.57) participated in EEG data collection for this data set \cite{Brunner:2008}. During recordings, subjects were sitting in front of a computer screen on which the cue-based BCI paradigm consisting of four motor imagery tasks was presented to them. Each subject participated in the experiment for two sessions on different days, henceforth referred to as \textit{session 1} and \textit{session 2}. Each of these sessions included six runs separated by short breaks, where each run consists of 48 trials (12 for each one of the four classes), yielding a total of 288 trials per session.

At the beginning of trials, a fixation cross was displayed on the black screen. After two seconds, a cue in the form of an arrow pointing either to the up, down, right or left corresponding to the four classes (i.e., tongue, feet, right hand or left hand imagination) appeared and stayed on the screen for 1.25 seconds. This instructed the subjects to perform the desired motor imagination task, with no feedback provided. Subjects were instructed to perform motor imagery until the fixation cross disappeared, which constituted a three seconds imagery time for data processing. Afterwards, a short break was displayed on the screen and the next trial began. The order of the cues (i.e., classes) across trials were randomized.

Twenty-two electrodes placed on the scalp according to the 10-20 system were used for EEG recordings at locations: Fz, FC3, FC1, FCz, FC2, FC4, C5, C3, C1, Cz, C2, C4, C6, CP3, CP1, CPz, CP2, CP4, P1, Pz, P2, POz. Data were referenced to the left mastoid, grounded to the right mastoid, sampled at 250 Hz, and filtered with 0.5--100 Hz band pass and a 50 Hz notch filter. We did not exclude any trials or perform any electrooculography (EOG) based artifact reduction.

\subsection{EEG Signal Processing Pipeline}

Three second imagery duration of trials results in a trial data matrix of 22 channels by 750 samples. Corresponding multi-class labels across trials are; tongue (class 1), both feet (class 2), right hand (class 3) and left hand (class 4). In binary hierarchical decoding, we analyzed the trials in the three one-versus-rest sub-problem levels intuitively as; (1) speech (class 1) versus motor (classes 2, 3, 4), (2) feet (class 2) versus hand (classes 3, 4), (3) right (class 3) versus left hand (class 4).

State-of-the-art discriminative spatial filtering of EEG in motor imagery paradigms highlights the common spatial patterns (CSP) algorithm \cite{Ramoser:2000,Blankertz:2008}, which aims to identify a discriminative basis for a multichannel signal recorded under different conditions, where signal representations maximally differ in variance between these conditions (i.e., classes). In a binary case, CSP algorithm aims to solve the problem:
\begin{equation}
\bm{w^*} = \underset{\bm{w}\in\mathbb{R}^N}{\text{argmax}} \Bigg\{ \frac{\bm{w}^T\Pi_1\bm{w}}{\bm{w}^T\Pi_2\bm{w}} \Bigg\},
\label{eq:cspobj}
\end{equation}
where $\Pi_1$ and $\Pi_2$ denote the $\mathit{N}\times\mathit{N}$ class covariance matrices of the data matrix with $\mathit{N}$ rows denoting the number of channels. Vector $\bm{w}$ indicates the discriminative spatial filter to be applied over channels. Eq. \ref{eq:cspobj} can be solved by the generalized eigenvalue problem: $\Pi_1\bm{w}=\lambda\Pi_2\bm{w}$, which has $\mathit{N}$ possible solutions. The eigenvector corresponding to the highest eigenvalue indicates a basis where variance of the class 1 data will be highest, and class 2 will be lowest. Vice versa for the lowest eigenvalue. Data pre-processing is usually performed by combining $K$ eigenvectors as pairs of smallest and largest eigenvalues obtained, forming $\bm{\mathit{W}}\in\mathbb{R}^{\mathit{N}\times K}$, and spatial filtering of the data with this matrix. Afterwards, $K$ dimensional features are calculated as the log-normalized signal variances across time-series of the CSP filtered data.

One prevalent extension of CSPs is band pass filtering of EEG in several frequency sub-bands before applying CSP, and concatenating the outputs of each sub-band specific CSP in one higher dimensional feature vector, which is known as the filter bank CSP (FBCSP) approach \cite{Ang:2008}. We exploit FBCSPs by band pass filtering EEG data for each trial and electrode in four frequency sub-bands known to be relevant for motor imagery; $\alpha$-band (8--12 Hz), $\beta_1$-band (12--16 Hz), $\beta_2$-band (16--22 Hz), and $\beta_3$-band (22--30 Hz). We used three pairs of CSP components ($K=6$) from each frequency sub-band, resulting in a 24 dimensional feature vector $\bm{x}_i$ at each trial $i$.

\renewcommand{\arraystretch}{1.15}
\begin{table*}
  \caption{Session 1 binary classification accuracies ($\%$) averaged over 5x5-fold cross validation repetitions. Values in parentheses indicate the standard deviations. Bold values indicate the highest mean accuracy across different feature learning methods.}
  \label{binary1resultstable}
  \centering
  \begin{tabular}{C{1cm} c | c c c c c c c c c | c}
    \toprule
    \multicolumn{2}{c|}{\textbf{Two Class - Session 1}} & \textbf{P1} & \textbf{P2} & \textbf{P3} & \textbf{P4} & \textbf{P5} & \textbf{P6} & \textbf{P7} & \textbf{P8} & \textbf{P9} & \textbf{Mean} \\
    \midrule
    \multirow{-0.5}{*}{\RotText{Speech vs Motor}} & CSP & 82.9 \tiny{(1.3)} & 79.5 \tiny{(1.5)} & 82.1 \tiny{(2.2)} & 69.7 \tiny{(1.7)} & 74.3 \tiny{(1.0)} & 69.7 \tiny{(1.0)} & 85.5 \tiny{(1.1)} & 91.3 \tiny{(0.5)} & 81.1 \tiny{(1.2)} & 79.5 \tiny{(1.2)} \\
    & FBCSP & 82.9 \tiny{(0.3)} & 80.9 \tiny{(0.8)} & 81.3 \tiny{(1.0)} & 78.8 \tiny{(1.9)} & \textbf{75.9} \textbf{\tiny{(1.2)}} & 75.2 \tiny{(2.0)} & 88.6 \tiny{(1.0)} & 94.3 \tiny{(0.4)} & \textbf{84.6} \textbf{\tiny{(0.3)}} & 82.5 \tiny{(0.9)} \\
    & $R^2$-Selection & 81.7 \tiny{(1.2)} & 78.4 \tiny{(1.4)} & 84.7 \tiny{(1.3)} & 78.3 \tiny{(1.5)} & 70.5 \tiny{(2.0)} & 74.0 \tiny{(3.5)} & 86.6 \tiny{(0.8)} & 92.2 \tiny{(0.5)} & 83.2 \tiny{(1.5)} & 81.0 \tiny{(1.5)} \\
    & SDA-Selection & 80.9 \tiny{(1.8)} & 80.6 \tiny{(2.2)} & 82.4 \tiny{(1.4)} & 77.2 \tiny{(3.1)} & 69.6 \tiny{(2.3)} & 73.2 \tiny{(2.8)} & 87.3 \tiny{(2.0)} & 93.2 \tiny{(1.2)} & 82.8 \tiny{(1.7)} & 80.8 \tiny{(2.0)} \\
    & mRMR-Selection & 82.2 \tiny{(1.3)} & 79.3 \tiny{(1.9)} & 82.4 \tiny{(1.7)} & 74.5 \tiny{(1.8)} & 69.9 \tiny{(2.0)} & 72.2 \tiny{(1.6)} & 87.4 \tiny{(2.4)} & 93.2 \tiny{(1.1)} & 82.6 \tiny{(2.9)} & 80.4 \tiny{(1.8)} \\
    & MMI-Selection & 82.5 \tiny{(1.5)} & 81.1 \tiny{(1.5)} & 84.1 \tiny{(1.5)} & \textbf{79.4} \textbf{\tiny{(3.0)}} & 71.2 \tiny{(2.5)} & 73.6 \tiny{(2.0)} & 86.2 \tiny{(1.7)} & 92.7 \tiny{(0.4)} & 84.3 \tiny{(1.2)} & 81.6 \tiny{(1.7)} \\
    & \cellc\textbf{MMI-LinT} & \cellc82.7 \tiny{(1.6)} & \cellc82.0 \tiny{(0.8)} & \cellc\textbf{85.2} \textbf{\tiny{(1.4)}} & \cellc79.2 \tiny{(1.5)} & \cellc75.2 \tiny{(1.3)} & \cellc\textbf{77.3} \textbf{\tiny{(2.0)}} & \cellc89.1 \tiny{(1.4)} & \cellc94.3 \tiny{(0.7)} & \cellc84.2 \tiny{(1.4)} & \cellc\textbf{83.2} \textbf{\tiny{(1.3)}} \\
    & \cellc\textbf{MMI-NonLinT} & \cellc\textbf{83.4} \tiny{(1.6)} & \cellc\textbf{82.3} \textbf{\tiny{(1.4)}} & \cellc83.8 \tiny{(0.5)} & \cellc77.5 \tiny{(1.1)} & \cellc75.6 \tiny{(1.0)} & \cellc76.5 \tiny{(0.9)} & \cellc\textbf{89.3} \textbf{\tiny{(1.6)}} & \cellc\textbf{94.6} \textbf{\tiny{(1.0)}} & \cellc84.2 \tiny{(1.9)} & \cellc83.0 \tiny{(1.2)} \\
    
    \midrule
    
    \multirow{-0.5}{*}{\RotText{Feet vs Hand}} & CSP & 91.0 \tiny{(1.6)} & \textbf{93.8} \textbf{\tiny{(0.9)}} & 90.7 \tiny{(1.5)} & 71.0 \tiny{(2.4)} & 69.8 \tiny{(2.0)} & 66.8 \tiny{(2.4)} & 95.3 \tiny{(0.5)} & 80.4 \tiny{(1.8)} & 69.9 \tiny{(2.4)} & 80.9 \tiny{(1.7)} \\
    & FBCSP & 90.4 \tiny{(1.6)} & 92.0 \tiny{(0.6)} & 92.6 \tiny{(1.6)} & 77.5 \tiny{(2.0)} & 72.7 \tiny{(2.1)} & 72.5 \tiny{(1.8)} & 96.4 \tiny{(1.2)} & 83.0 \tiny{(1.3)} & 73.8 \tiny{(3.0)} & 83.4 \tiny{(1.6)} \\
    & $R^2$-Selection & 90.5 \tiny{(1.0)} & 90.3 \tiny{(2.1)} & 93.8 \tiny{(1.2)} & 75.3 \tiny{(2.2)} & 68.4 \tiny{(4.0)} & 67.0 \tiny{(2.8)} & \textbf{97.5 \tiny{(0.6)}} & 82.2 \tiny{(2.4)} & 74.6 \tiny{(2.6)} & 82.1 \tiny{(1.9)} \\
    & SDA-Selection & 91.5 \tiny{(0.6)} & 90.2 \tiny{(1.8)} & 94.8 \tiny{(0.6)} & 73.9 \tiny{(1.7)} & 71.0 \tiny{(3.5)} & 74.3 \tiny{(2.1)} & 96.5 \tiny{(0.8)} & 81.2 \tiny{(1.2)} & 74.3 \tiny{(1.6)} & 83.0 \tiny{(1.5)} \\
    & mRMR-Selection & 90.7 \tiny{(1.9)} & 90.7 \tiny{(1.0)} & 94.0 \tiny{(1.3)} & 71.0 \tiny{(2.2)} & 69.6 \tiny{(4.2)} & 70.3 \tiny{(3.9)} & 97.3 \tiny{(1.2)} & 78.7 \tiny{(2.3)} & 73.6 \tiny{(0.9)} & 81.7 \tiny{(2.1)} \\
    & MMI-Selection & 90.8 \tiny{(1.2)} & 91.3 \tiny{(2.0)} & \textbf{95.7} \textbf{\tiny{(1.0)}} & 72.8 \tiny{(3.0)} & 70.8 \tiny{(1.9)} & 71.3 \tiny{(2.6)} & 96.8 \tiny{(1.1)} & \textbf{83.7} \textbf{\tiny{(2.9)}} & 73.8 \tiny{(2.6)} & 83.0 \tiny{(2.0)} \\
    & \cellc\textbf{MMI-LinT} & \cellc\textbf{93.1} \textbf{\tiny{(1.5)}} & \cellc92.8 \tiny{(0.7)} & \cellc95.0 \tiny{(0.9)} & \cellc78.1 \tiny{(3.0)} & \cellc\textbf{73.8} \textbf{\tiny{(2.8)}} & \cellc\textbf{75.0} \textbf{\tiny{(1.5)}} & \cellc96.7 \tiny{(0.3)} & \cellc81.1 \tiny{(1.9)} & \cellc72.7 \tiny{(1.8)} & \cellc84.2 \tiny{(1.6)} \\
    & \cellc\textbf{MMI-NonLinT} & \cellc92.5 \tiny{(1.0)} & \cellc92.5 \tiny{(0.8)} & \cellc94.5 \tiny{(1.1)} & \cellc\textbf{78.7} \textbf{\tiny{(2.1)}} & \cellc73.6 \tiny{(1.2)} & \cellc74.9 \tiny{(3.1)} & \cellc97.0 \tiny{(0.5)} & \cellc81.5 \tiny{(0.8)} & \cellc\textbf{75.0} \textbf{\tiny{(2.5)}} & \cellc\textbf{84.4} \textbf{\tiny{(1.4)}} \\
    
    \midrule
    
    \multirow{-0.5}{*}{\RotText{Right vs Left}} & CSP & 82.2 \tiny{(1.5)} & \textbf{65.2} \textbf{\tiny{(4.9)}} & 94.0 \tiny{(1.4)} & 73.1 \tiny{(4.1)} & 59.1 \tiny{(2.0)} & 64.8 \tiny{(1.6)} & 75.8 \tiny{(3.2)} & \textbf{96.6} \textbf{\tiny{(0.9)}} & 70.8 \tiny{(2.4)} & 75.7 \tiny{(2.4)} \\
    & FBCSP & 84.7 \tiny{(0.4)} & 60.5 \tiny{(3.1)} & 93.4 \tiny{(1.7)} & 74.8 \tiny{(2.2)} & 70.1 \tiny{(2.1)} & 63.4 \tiny{(1.7)} & 73.1 \tiny{(3.5)} & 94.8 \tiny{(1.8)} & 71.6 \tiny{(1.5)} & 76.2 \tiny{(2.0)} \\
    & $R^2$-Selection & 83.8 \tiny{(1.3)} & 59.3 \tiny{(4.9)} & 93.6 \tiny{(1.8)} & 70.0 \tiny{(2.8)} & 68.7 \tiny{(4.8)} & 59.0 \tiny{(2.6)} & 75.2 \tiny{(1.5)} & 94.1 \tiny{(1.3)} & 68.4 \tiny{(2.0)} & 74.6 \tiny{(2.5)} \\
    & SDA-Selection & 84.5 \tiny{(2.4)} & 58.7 \tiny{(2.5)} & \textbf{94.1} \textbf{\tiny{(1.2)}} & 70.6 \tiny{(2.9)} & 69.8 \tiny{(2.1)} & 63.6 \tiny{(1.8)} & 72.9 \tiny{(3.9)} & 94.0 \tiny{(1.8)} & 70.8 \tiny{(3.9)} & 75.4 \tiny{(2.5)} \\
    & mRMR-Selection & 82.7 \tiny{(2.1)} & 54.3 \tiny{(3.2)} & 90.4 \tiny{(1.3)} & 70.1 \tiny{(5.1)} & 67.7 \tiny{(2.7)} & 60.4 \tiny{(4.1)} & 71.6 \tiny{(3.1)} & 89.7 \tiny{(1.7)} & 66.8 \tiny{(3.3)} & 72.6 \tiny{(2.9)} \\
    & MMI-Selection & \textbf{87.6} \textbf{\tiny{(1.9)}} & 63.0 \tiny{(4.8)} & 94.0 \tiny{(2.6)} & 70.8 \tiny{(1.3)} & \textbf{72.9} \textbf{\tiny{(2.0)}} & 63.7 \tiny{(2.2)} & 73.1 \tiny{(0.7)} & 95.2 \tiny{(0.9)} & \textbf{72.5} \textbf{\tiny{(2.8)}} & 76.9 \tiny{(2.1)} \\
    & \cellc\textbf{MMI-LinT} & \cellc86.3 \tiny{(1.8)} & \cellc61.3 \tiny{(3.4)} & \cellc93.8 \tiny{(1.5)} & \cellc\textbf{75.5} \textbf{\tiny{(2.4)}} & \cellc72.7 \tiny{(5.4)} & \cellc65.8 \tiny{(4.6)} & \cellc\textbf{76.3} \textbf{\tiny{(2.3)}} & \cellc92.5 \tiny{(1.5)} & \cellc69.3 \tiny{(1.7)} & \cellc77.0 \tiny{(2.7)} \\
    & \cellc\textbf{MMI-NonLinT} & \cellc86.2 \tiny{(2.6)} & \cellc61.9 \tiny{(3.3)} & \cellc93.8 \tiny{(2.7)} & \cellc75.2 \tiny{(4.5)} & \cellc72.5 \tiny{(4.3)} & \cellc\textbf{66.2} \textbf{\tiny{(2.0)}} & \cellc75.0 \tiny{(2.5)} & \cellc92.5 \tiny{(1.4)} & \cellc71.1 \tiny{(4.7)} & \cellc\textbf{77.1} \textbf{\tiny{(3.1)}} \\
    \bottomrule
  \end{tabular}
\end{table*}

\subsection{Feature Learning Frameworks and Classification}

We assess our approach in comparison to using raw CSP or FBCSP features as a methodological baseline, coefficient of determination based statistical feature selection ($R^2$-Selection) from the FBCSP feature vectors \cite{Muller:2004}, stepwise discriminant analysis based selection of features (SDA-Selection) as explored by \cite{Krusienski:2008}, conventional maximum mutual information based feature ranking and selection (MMI-Selection) for dimensionality reduction \cite{Ang:2012}, and another mutual information driven approach of minimum Redundancy Maximum Relevance feature selection (mRMR-Selection) \cite{Peng:2005} as explored by \cite{Muhl:2014}. Implementations of the methods are presented below.
\begin{enumerate}
    \item \textit{CSP}: A single 8-30 Hz band pass filter was applied. No dimensionality reduction was performed for the feature vector, resulting in $d_x$ = $d_y$ = 6.
    \item \textit{FBCSP}: No dimensionality reduction was performed. Likelihood density estimations were performed with feature vector dimensions $d_x$ = $d_y$ = 24.
    \item \textit{$R^2$-Selection}: $R^2$ statistics based feature ranking and selection \cite{Muller:2004} was performed for the FBCSP feature vector $d_x$ = 24 to reduce to $d_y$ = 6.
    \item \textit{SDA-Selection}: SDA utilizes a combination of forward and backward statistical significance based selection steps: (1) weighting the training features using ordinary least squares regression (i.e., Fisher's linear discriminant) to predict their labels, (2) starting with an empty set of selected features, the most significant input feature ($p<$ 0.05) in prediction is selected and added to the discriminant function, (3) a backward step to remove the least significant input feature from the discriminant function ($p>$ 0.05), (4) repeat until no more features satisfy the forward or backward criteria. In our implementations, the number of selected features was resulting in a maximum dimensionality of six ($d_y\le$ 6). Algorithm was implemented based on \cite{Jennrich:1977,Draper:1981}.
    \item \textit{mRMR-Selection}: mRMR algorithm relies on a mutual information based minimal-redundancy-maximal-relevance criterion between features and labels for incremental feature selection. For mutual information computations, the original algorithm suggests a priori discretizing the continuous feature variables. Hence, we discretized features in three states based on the mean and standard deviation across samples \cite{Peng:2005}. Number of selected features was chosen as 6, in consistency with the other methods ($d_x$ = 24, $d_y$ = 6).
    \item \textit{MMI-Selection}: Based on maximum mutual information ranking, selections are also performed in pairs by nature of CSP (i.e., high/low eigenvector projection pair of any ranking based selected feature is also selected) \cite{Ang:2012}. We investigated MMI based selection of either 2, 4, or 6 features ($d_x$ = 24, $d_y\in\{2,4,6\}$). Only highest decoding accuracies across these three are reported. Feature selection dimensionalities higher than 6 were not considered due to lower accuracies since they tend to fail in kernel density estimations and prone to overfitting.
    \item \textit{MMI-LinT}: Dimensionality reduction of the FBCSP feature vector ($d_x$ = 24) to two dimensions ($d_y$ = 2) is performed based on Section \ref{sec:mmilint}. Number of training epochs were 20, with 0.01 gradient step size, and the momentum parameter for optimization being 0.9.
    \item \textit{MMI-NonLinT}: Dimensionality reduction of the FBCSP feature vector ($d_x$ = 24) to two dimensions ($d_y$ = 2) is performed based on Section \ref{sec:mminonlint}. Number of nodes in the hidden layer were chosen to be 30, the number of training epochs were 20, gradient step size 0.01, and the momentum parameter for optimization being 0.9.
\end{enumerate}

To evaluate Eq.~\ref{eq:decisioncriteria3} for multi-class hierarchical decoding, class priors were assumed to be uniform, and the class conditional densities were derived by multivariate Gaussian kernel density estimation with bandwidth sizes determined by Silverman’s rule \cite{Silverman:1986}. Analogously, we demonstrate the feasibility of our approach for binary decoding level-wise. Here, classification was based on MAP estimation over two class labels using Gaussian kernel density estimation of likelihoods, which can be interpreted as the kernel density classifier. In comparison to inheriting assumptions (e.g., Gaussianity of likelihoods for linear discriminant analysis which are widely favored for BCIs \cite{Garrett:2003,Muller:2003}), the kernel density classifier is not parametrically restricted besides the innate choice of kernels. However, the vulnerability may arise from unstability in high dimensional regions where there is little training data.

\subsection{Binary Classification Results}

We report binary decoding accuracies at hierarchical sub-problems as: (1) speech versus motor, (2) feet versus hand, (3) right hand versus left hand (see Table~\ref{binary1resultstable}). To demonstrate the feasibility of our approach in binary decoding, which we also previously studied in \cite{Ozdenizci:2017b}, we only present these results on the session 1 data sets using 5-fold cross validation, which were repeated 5 times. Across most of the subjects and on average, MMI-LinT and MMI-NonLinT outperforms the baseline and feature selection frameworks in decoding. 

Paired t-tests for statistical significance of performance ($p<$ 0.05) between the best performing feature learning approach and the other methods at each level are performed. For speech versus motor, MMI-LinT revealed significant difference from CSP ($p$ = 0.006), $R^2$- ($p$ = 0.002), SDA- and mRMR-Selection ($p$ = 0.001), as well as MMI-Selection ($p$ = 0.02). However no significant difference to FBCSP ($p$ = 0.18), and MMI-NonLinT ($p$ = 0.42) was observed. For feet versus hand, MMI-NonLinT revealed significant difference from CSP ($p$ = 0.009), FBCSP and SDA-Selection ($p$ = 0.02), $R^2$- ($p$ = 0.04), and mRMR-Selection ($p$ = 0.01), but not MMI-Selection ($p$ = 0.10), or MMI-LinT ($p$ = 0.51). For right versus left, MMI-NonLinT revealed significant differences to $R^2$- and SDA-Selection ($p$ = 0.02), as well as mRMR-Selection ($p$ = 0.001).

\renewcommand{\arraystretch}{1.15}
\begin{table*}
  \caption{Four class classification accuracies ($\%$) averaged over 5x5-fold cross validation repetitions in within session analyses, and averaged over two session-to-session decoding accuracies in across sessions analyses (i.e., model training on session 1 and testing on session 2, and vice versa). Values in parentheses indicate the standard deviations. Bold values indicate the highest mean accuracies across different feature learning methods.}
  \label{multiresults}
  \centering
  \begin{tabular}{C{1cm} c | c c c c c c c c c | c}
    \toprule
    \multicolumn{2}{c|}{\textbf{Four Class Decoding}} & \textbf{P1} & \textbf{P2} & \textbf{P3} & \textbf{P4} & \textbf{P5} & \textbf{P6} & \textbf{P7} & \textbf{P8} & \textbf{P9} & \textbf{Mean} \\
    \midrule
    \multirow{-0.5}{*}{\RotText{Within Session 1 (5x5 Fold CV)}} & CSP &  70.2  \tiny{(0.5)} &  62.0  \tiny{(1.8)} &  76.4  \tiny{(1.8)} &  48.2  \tiny{(2.5)} &  40.3  \tiny{(1.3)} &  42.7  \tiny{(2.0)} &  74.0  \tiny{(1.2)} &  76.9  \tiny{(1.7)} &  53.7  \tiny{(0.5)} &  60.4  \tiny{(1.4)} \\
    & FBCSP &  68.9  \tiny{(2.0)} &  59.6  \tiny{(2.1)}&  75.3  \tiny{(2.1)} &  55.2  \tiny{(2.5)} &  46.1  \tiny{(2.4)} &  45.4  \tiny{(1.8)} &  73.8  \tiny{(2.5)} &  79.5  \tiny{(2.2)} &  56.6  \tiny{(1.3)} &  62.2  \tiny{(2.1)} \\
    & $R^2$-Selection & 67.6 \tiny{(2.1)} & 58.6 \tiny{(1.6)} & 78.5 \tiny{(1.4)} & 49.7 \tiny{(2.0)} & 43.0 \tiny{(3.7)} & 43.0 \tiny{(1.8)} & 72.2 \tiny{(1.2)} & 76.8 \tiny{(2.0)} & 56.1 \tiny{(1.4)} & 60.6 \tiny{(1.9)} \\
    & SDA-Selection &  70.3  \tiny{(2.3)} &  59.5  \tiny{(2.5)} &  77.0  \tiny{(1.0)} &  52.1  \tiny{(1.8)} &  44.8  \tiny{(1.8)} &  45.3  \tiny{(1.2)} &  73.3  \tiny{(1.4)} &  80.4  \tiny{(0.8)} &  56.6  \tiny{(1.7)} &  62.1  \tiny{(1.5)} \\
    & mRMR-Selection & 69.7 \tiny{(3.7)} & 56.8 \tiny{(1.4)} & 74.3 \tiny{(2.9)} & 50.8 \tiny{(4.8)} & 42.3 \tiny{(3.4)} & 43.8 \tiny{(2.8)} & 73.9 \tiny{(4.7)} & 73.4 \tiny{(1.5)} & 53.8 \tiny{(3.1)} & 59.8 \tiny{(3.1)} \\
    & MMI-Selection &  69.3  \tiny{(3.5)} &  59.7  \tiny{(2.3)} &  77.9  \tiny{(2.4)} &  52.8  \tiny{(1.7)} &  44.5  \tiny{(3.8)} &  44.5  \tiny{(2.2)} &  73.6  \tiny{(1.7)} &  77.2  \tiny{(0.9)} &  57.4  \tiny{(1.9)} &  61.8  \tiny{(2.2)} \\
    & \cellc\textbf{MMI-LinT} & \cellc \textbf{74.2}  \textbf{\tiny{(1.8)}} & \cellc \textbf{64.1}  \textbf{\tiny{(2.3)}} & \cellc 79.6  \tiny{(2.2)} & \cellc 55.2  \tiny{(2.0)} & \cellc 48.4  \tiny{(2.2)} & \cellc \textbf{50.5}  \textbf{\tiny{(2.5)}} & \cellc \textbf{77.1}  \textbf{\tiny{(2.4)}} & \cellc \textbf{79.2}  \textbf{\tiny{(0.8)}} & \cellc 57.7  \tiny{(2.4)} & \cellc \textbf{65.1}  \textbf{\tiny{(2.0)}} \\
    & \cellc\textbf{MMI-NonLinT} & \cellc 72.2  \tiny{(2.0)} & \cellc 62.9  \tiny{(1.3)} & \cellc \textbf{79.7}  \textbf{\tiny{(1.7)}} & \cellc \textbf{55.3}  \textbf{\tiny{(3.7)}} & \cellc \textbf{48.8}  \textbf{\tiny{(3.0)}} & \cellc 48.8  \tiny{(4.4)} & \cellc 76.6  \tiny{(1.5)} & \cellc 78.9  \tiny{(1.9)} & \cellc \textbf{57.9}  \textbf{\tiny{(1.8)}} & \cellc 64.5  \tiny{(2.3)} \\
    
    \midrule
    
    \multirow{-0.5}{*}{\RotText{Within Session 2 (5x5 Fold CV)}} & CSP &  66.8  \tiny{(1.6)} &  57.2  \tiny{(0.3)} &  69.3  \tiny{(1.1)} &  59.7  \tiny{(1.9)} &  40.8  \tiny{(1.1)} &  38.0  \tiny{(3.1)} &  81.3  \tiny{(0.8)} &  \textbf{72.9}  \textbf{\tiny{(2.4)}} &  83.3  \tiny{(1.1)} &  63.2  \tiny{(1.4)} \\
    & FBCSP &  70.2  \tiny{(1.4)} &  55.8  \tiny{(0.5)} &  75.1  \tiny{(2.4)} &  60.6  \tiny{(3.1)} &  39.4  \tiny{(1.1)} &  38.8  \tiny{(1.1)} &  81.7  \tiny{(1.5)} &  70.4  \tiny{(1.3)} &  82.9  \tiny{(1.2)} &  63.8  \tiny{(1.5)} \\
    & $R^2$-Selection & 69.5 \tiny{(1.5)} & 54.3 \tiny{(1.3)} & 78.3 \tiny{(1.8)} & 57.5 \tiny{(1.9)} & 41.8 \tiny{(3.1)} & 40.8 \tiny{(3.0)} & 80.0 \tiny{(1.7)} & 66.1 \tiny{(1.6)} & 82.5 \tiny{(1.6)} & 63.4 \tiny{(1.9)} \\
    & SDA-Selection &  70.9  \tiny{(1.3)} &  56.5  \tiny{(2.0)} &  77.7  \tiny{(1.5)} &  57.4  \tiny{(2.3)} &  42.7  \tiny{(2.3)} &  40.6  \tiny{(4.0)} &  80.9  \tiny{(2.1)} &  67.8  \tiny{(2.1)} &  84.7  \tiny{(1.4)} &  64.3  \tiny{(2.1)} \\
    & mRMR-Selection & 68.1 \tiny{(3.1)} & 52.5 \tiny{(1.8)} & 76.5 \tiny{(2.7)} & 56.8 \tiny{(2.8)} & 40.4 \tiny{(4.1)} & 42.2 \tiny{(4.1)} & 76.2 \tiny{(1.4)} & 63.4 \tiny{(3.2)} & 82.8 \tiny{(1.9)} & 62.1 \tiny{(2.7)} \\
    & MMI-Selection &  70.9  \tiny{(2.9)} &  55.6  \tiny{(0.8)} &  78.7  \tiny{(2.4)} &  58.6  \tiny{(4.1)} &  43.5  \tiny{(2.9)} &  41.9  \tiny{(2.1)} &  \textbf{82.3}  \textbf{\tiny{(1.1)}} &  68.4  \tiny{(1.6)} &  \textbf{86.2}  \textbf{\tiny{(2.0)}} &  65.1  \tiny{(2.2)} \\
    & \cellc\textbf{MMI-LinT} & \cellc \textbf{72.4}  \textbf{\tiny{(2.4)}} & \cellc 60.9  \tiny{(1.8)} & \cellc 79.6  \tiny{(2.3)} & \cellc \textbf{65.9}  \textbf{\tiny{(2.2)}} & \cellc 44.2  \tiny{(2.7)} & \cellc 48.4  \tiny{(3.3)} & \cellc 81.1  \tiny{(1.6)} & \cellc 72.2  \tiny{(3.5)} & \cellc 84.4  \tiny{(1.5)} & \cellc 67.6  \tiny{(2.3)} \\
    & \cellc\textbf{MMI-NonLinT} & \cellc 72.0  \tiny{(1.4)} & \cellc \textbf{61.4}  \textbf{\tiny{(1.7)}} & \cellc \textbf{80.4}  \textbf{\tiny{(1.2)}} & \cellc 65.0  \tiny{(2.6)} & \cellc \textbf{44.9}  \textbf{\tiny{(1.7)}} & \cellc \textbf{49.4}  \textbf{\tiny{(2.2)}} & \cellc 82.2  \tiny{(1.2)} & \cellc 71.8  \tiny{(1.3)} & \cellc 83.6  \tiny{(1.3)} & \cellc \textbf{67.8}  \textbf{\tiny{(1.6)}} \\
    
    \midrule
    
    \multirow{-0.5}{*}{\RotText{Across Sessions}} & CSP &  65.6 \tiny{(3.4)}  &  \textbf{49.5} \textbf{\tiny{(3.7)}}  &  66.1 \tiny{(3.2)}  &  44.2 \tiny{(6.1)}  &  28.7 \tiny{(2.7)}  &  41.0 \tiny{(1.9)} &  63.2 \tiny{(1.0)} &  66.0 \tiny{(2.9)} &  \textbf{59.7} \textbf{\tiny{(9.3)}}  &  53.7 \tiny{(3.8)} \\
    & FBCSP &  67.9 \tiny{(0.3)} &  47.7 \tiny{(1.7)} &  69.2 \tiny{(1.2)} &  \textbf{48.8} \textbf{\tiny{(13.5)}} &  37.1 \tiny{(3.9)} &  38.2 \tiny{(1.5)} & 71.0 \tiny{(2.2)} &  \textbf{67.7} \textbf{\tiny{(3.4)}} & 56.0 \tiny{(4.4)} &  55.9 \tiny{(3.5)} \\
    & $R^2$-Selection & 62.0 \tiny{(0.7)} & 48.2 \tiny{(1.5)} & 68.4 \tiny{(8.8)} & 45.5 \tiny{(1.5)} & 35.2 \tiny{(0.3)} & 37.8 \tiny{(1.5)} & 64.6 \tiny{(9.8)} & 64.2 \tiny{(6.9)} & 56.4 \tiny{(6.1)} & 53.5 \tiny{(4.1)} \\
    & SDA-Selection &  66.7 \tiny{(2.9)} &  40.8 \tiny{(15.9)} &  68.2 \tiny{(4.2)} &  43.2 \tiny{(3.2)} &  36.6 \tiny{(1.7)} &  44.6 \tiny{(3.7)} &  70.1 \tiny{(5.9)} &  60.2 \tiny{(0.3)} &  57.6 \tiny{(5.4)} &  54.2 \tiny{(4.8)} \\
    & mRMR-Selection & 67.0 \tiny{(0.5)} & 43.8 \tiny{(4.9)} & 68.9 \tiny{(2.7)} & 46.2 \tiny{(5.4)} & 33.3 \tiny{(4.4)} & 39.2 \tiny{(1.0)} & 61.6 \tiny{(5.7)} & 57.6 \tiny{(0.5)} & 57.6 \tiny{(6.4)} & 52.8 \tiny{(3.5)} \\
    & MMI-Selection &  65.5 \tiny{(2.7)} &  49.1 \tiny{(2.7)} &  69.3 \tiny{(7.6)} &  46.2 \tiny{(3.4)} &  38.0 \tiny{(2.9)} &  40.1 \tiny{(0.7)} &  67.7 \tiny{(5.9)} &  65.6 \tiny{(4.9)} &  57.3 \tiny{(2.9)} &  55.4 \tiny{(3.7)} \\
    & \cellc\textbf{MMI-LinT} & \cellc \textbf{68.8} \textbf{\tiny{(5.4)}}  & \cellc 47.4 \tiny{(4.7)}  & \cellc \textbf{74.3} \textbf{\tiny{(2.9)}} & \cellc 44.8 \tiny{(2.5)} & \cellc 37.9 \tiny{(0.5)} & \cellc 46.2 \tiny{(4.4)} & \cellc 71.0 \tiny{(10.6)} & \cellc 67.4 \tiny{(3.9)} & \cellc 56.6 \tiny{(3.9)} & \cellc \textbf{57.1} \textbf{\tiny{(4.3)}} \\
    & \cellc\textbf{MMI-NonLinT} & \cellc 67.9 \tiny{(4.7)} & \cellc 45.8 \tiny{(11.8)} & \cellc 73.6 \tiny{(0.0)} & \cellc 46.2 \tiny{(2.5)} & \cellc \textbf{38.2} \textbf{\tiny{(3.4)}} & \cellc \textbf{47.2} \textbf{\tiny{(2.5)}} & \cellc \textbf{72.4} \textbf{\tiny{(6.6)}} & \cellc 63.4 \tiny{(3.7)} & \cellc 58.2 \tiny{(4.7)} & \cellc 56.9 \tiny{(4.4)} \\
    
    \bottomrule
  \end{tabular}
\end{table*}

\subsection{Multi-Class Classification Results}

\begin{figure*}
	\vspace{-0.1cm}
	\centering
	\subfigure{\includegraphics[width=0.33\textwidth]{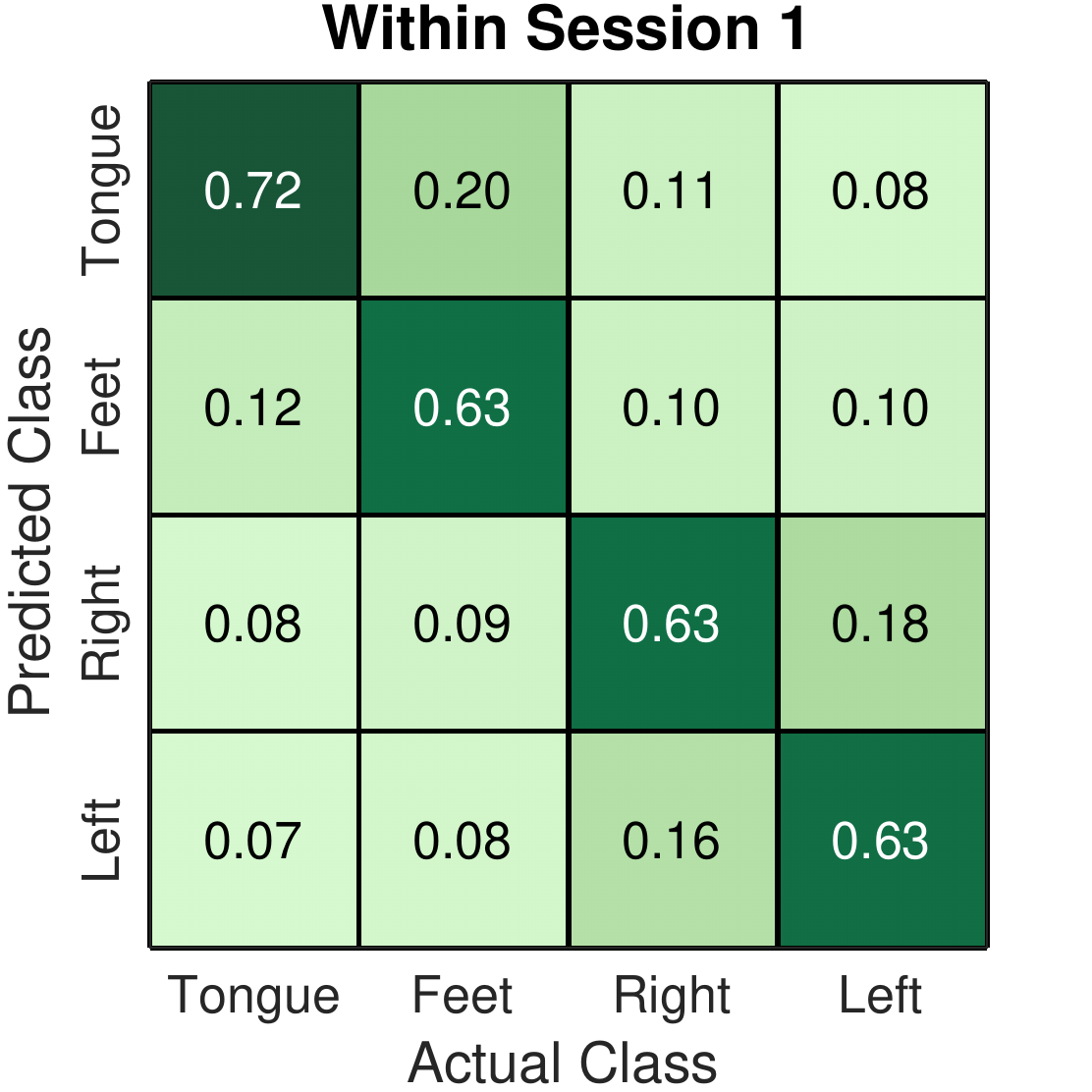}}\hspace{-0.2cm}
	\subfigure{\includegraphics[width=0.33\textwidth]{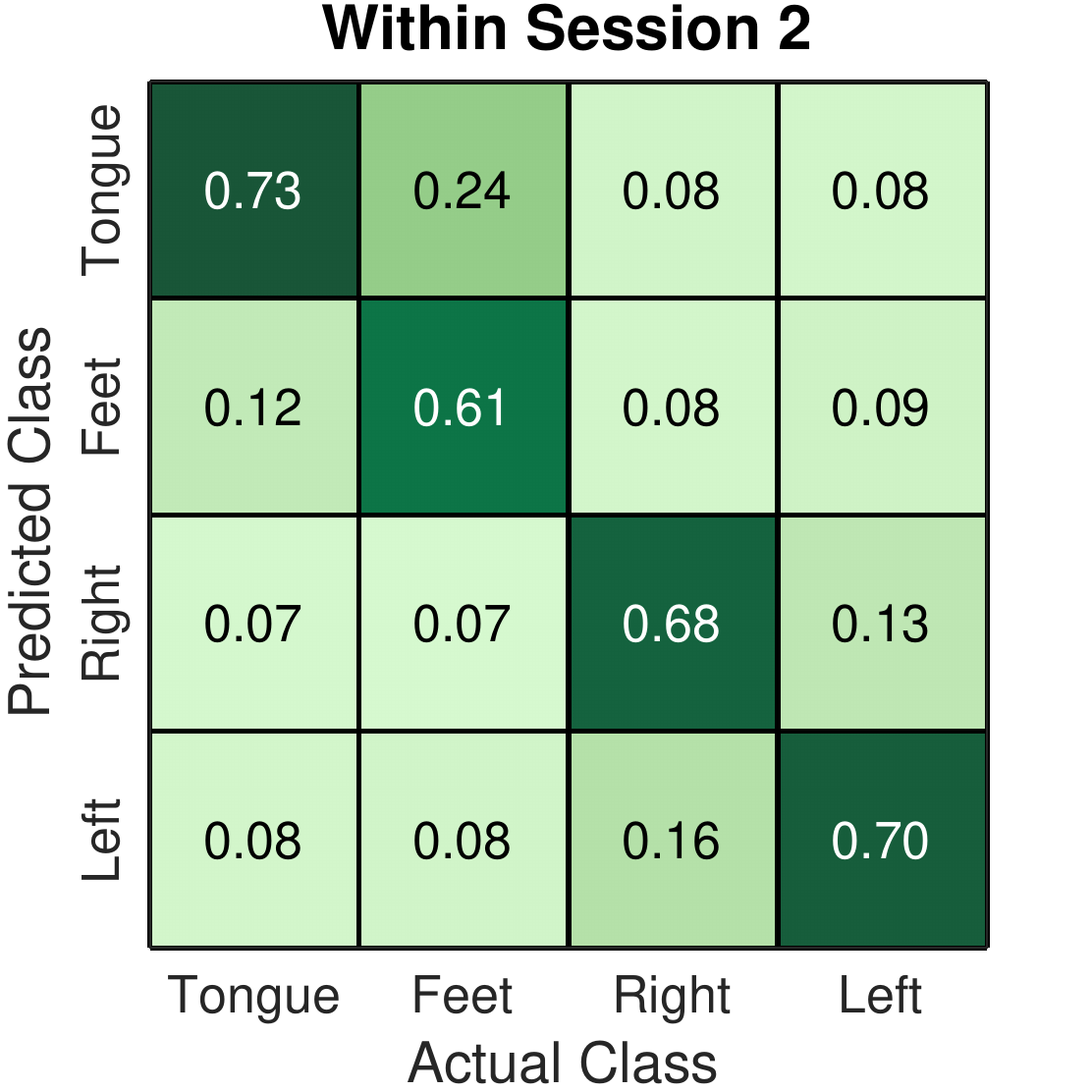}}\hspace{-0.2cm}
	\subfigure{\includegraphics[width=0.33\textwidth]{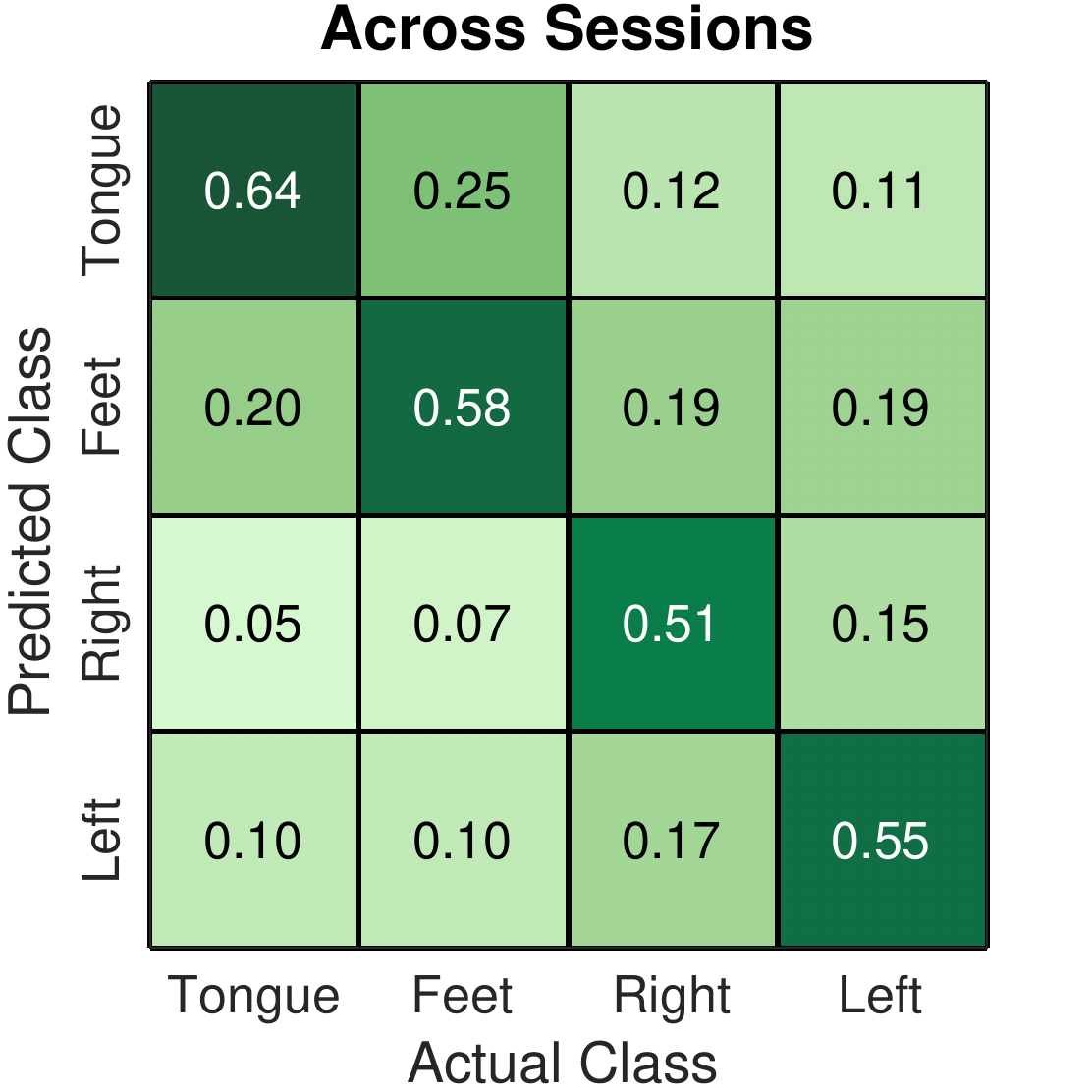}}
	\vspace{-0.15cm}
    \caption{Predicted versus actual class decoding accuracies shown in $[0,1]$ range, averaged across subjects and cross validation repetitions, for the four class problem. Results are computed for the feature learning protocols that produced the highest mean accuracies in Table~\ref{multiresults}; within session 1 with MMI-LinT ($65.1\%$), within session 2 with MMI-NonLinT ($67.8\%$), across sessions with MMI-LinT ($57.1\%$). All values are rounded to the nearest hundredth.}
	\label{fig:confMatrices}
\end{figure*}

Multi-class classification based on the hierarchical decoding approach was performed as: (1) 5x5-fold cross validation on session 1 data, (2) 5x5-fold cross validation on session 2 data, (3) two across sessions analyses (i.e., training on session 1 and testing on session 2 data, and vice versa). Our results demonstrate that MMI-LinT and MMI-NonLinT outperforms other methods in multi-class decoding, where the problem is highly prone to overfitting of high-dimensional features or heuristic feature selection algorithms (see Table~\ref{multiresults}). Highest mean decoding accuracy for within session 1 and across sessions analyses are observed with MMI-LinT ($65.1\%$ and $57.1\%$), and for within session 2 analyses is observed with MMI-NonLinT ($67.8\%$). Figure~\ref{fig:confMatrices} depicts the four class decoding confusion matrices between actual and predicted class labels of these best performing feature learning approaches.

Paired t-tests between the proposed and the other methods for within and across sessions analyses are performed. For within session 1, MMI-LinT revealed significant difference from FBCSP with $p$ = 0.004, as well as all the other methods with $p$ = 0.001. Similarly MMI-NonLinT revealed significant difference from FBCSP with $p$ = 0.003, SDA-Selection with $p$ = 0.002, as well as all the other methods with $p$ = 0.001. For within session 2, MMI-LinT revealed significant difference from CSP and SDA-Selection ($p$ = 0.01), $R^2$-Selection and FBCSP ($p$ = 0.004), mRMR- ($p$ = 0.001) and MMI-Selection ($p$ = 0.04). Likewise MMI-NonLinT revealed significant difference from CSP ($p$ = 0.01), FBCSP ($p$ = 0.005), $R^2$- and mRMR-Selection ($p$ = 0.001), SDA- ($p$ = 0.009) and MMI-Selection ($p$ = 0.03). For across sessions, MMI-LinT versus $R^2$-, SDA- and mRMR-Selection ($p$ = 0.01) showed significant differences. Similarly MMI-NonLinT versus $R^2$- ($p$ = 0.03), mRMR- ($p$ = 0.009) and SDA-Selection ($p$ = 0.001) revealed significant differences. The other paired comparisons with respect our methods did not show significant differences in across sessions analyses ($p>$ 0.05).

\begin{figure*}
	\centering
	\subfigure{\includegraphics[width=0.33\textwidth]{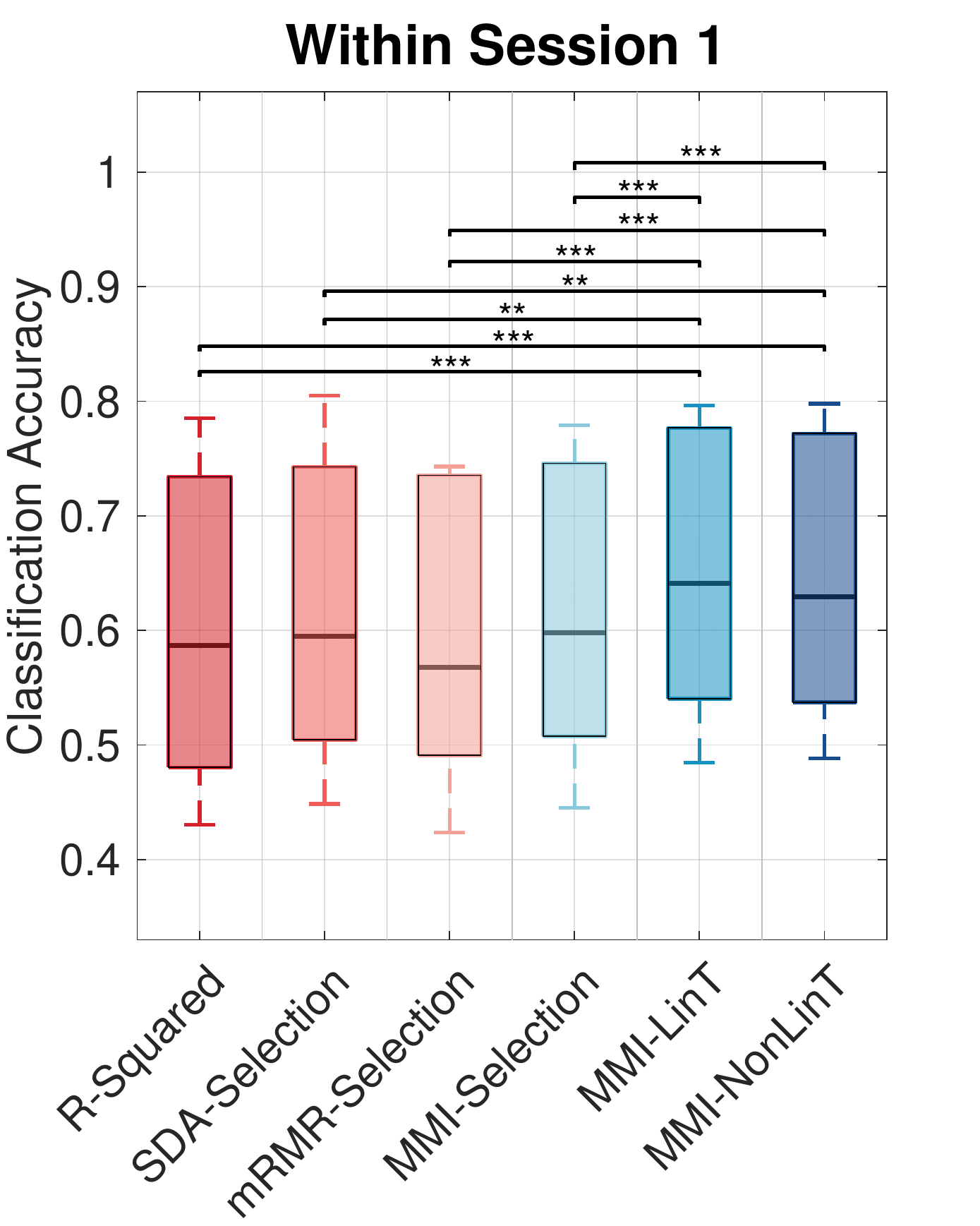}}\hspace{-0.2cm}
	\subfigure{\includegraphics[width=0.33\textwidth]{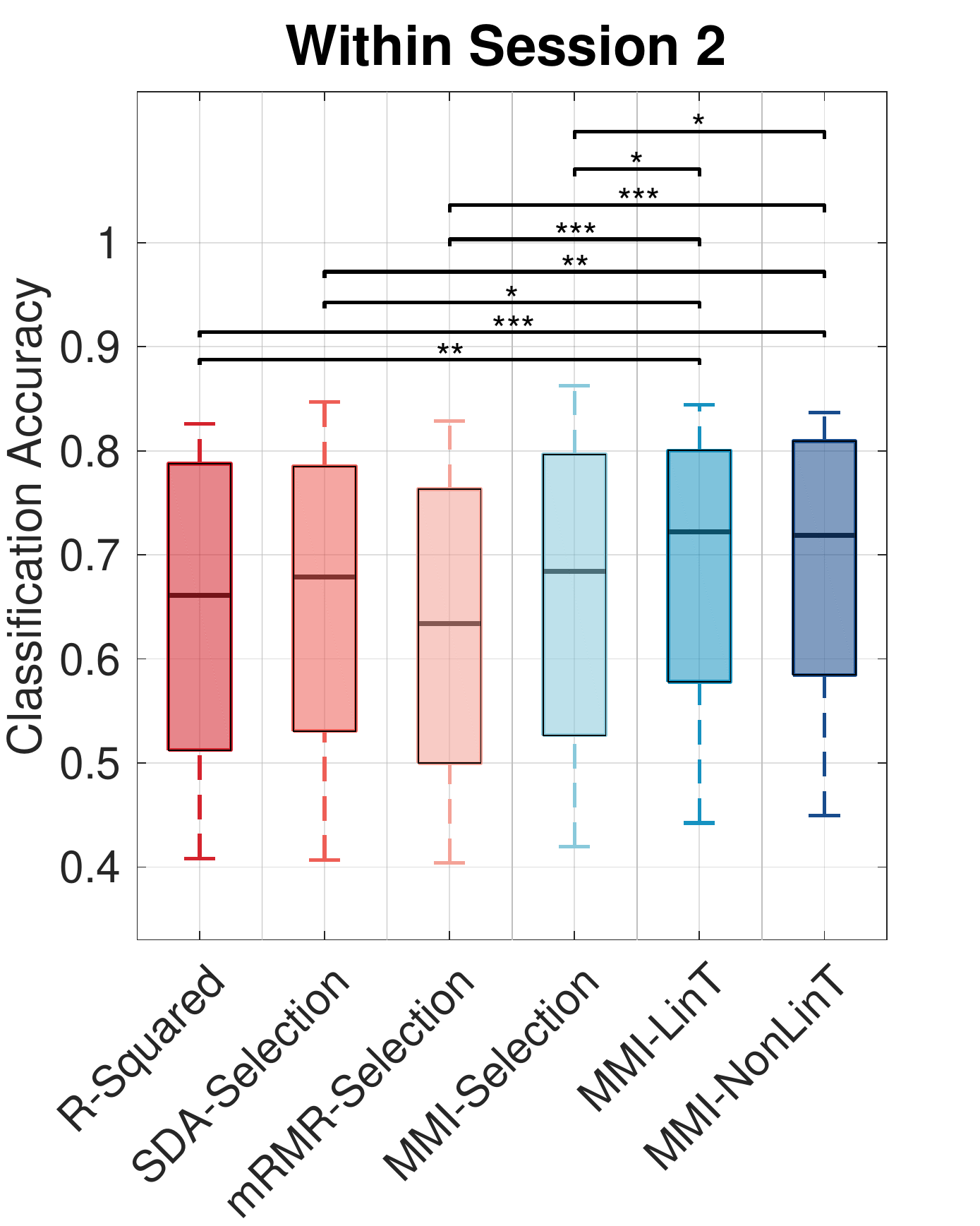}}\hspace{-0.2cm}
	\subfigure{\includegraphics[width=0.33\textwidth]{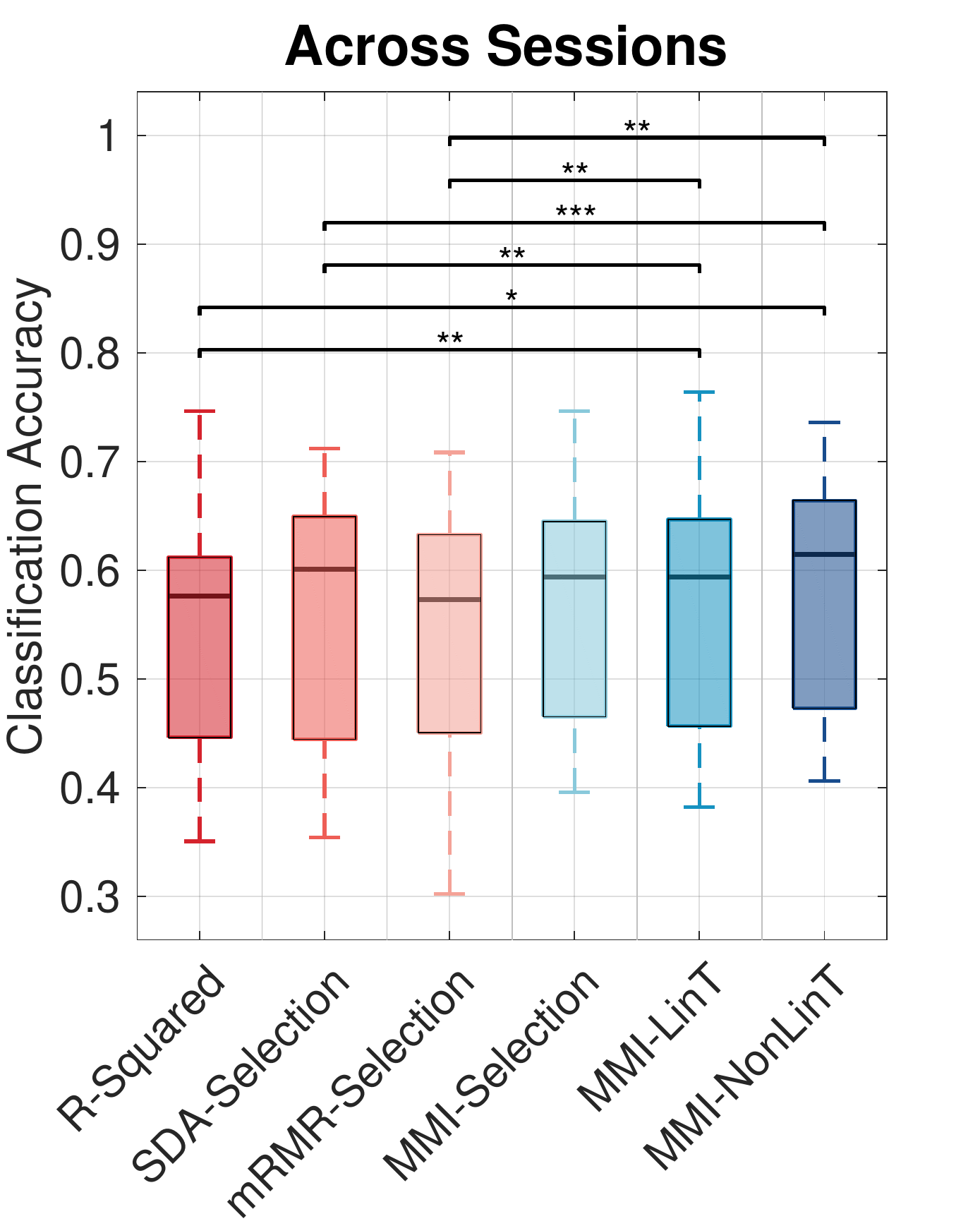}}
	\vspace{-0.15cm}
    \caption{Accuracies in $[0,1]$ range across subjects for the dimensionality reduction methods in Table~\ref{multiresults}. Central line mark represents the median across subjects. The upper and lower edges of the box represent the first and third quartiles. Upper and lower ends of the dashed lines represent the extreme data points. Starred marks indicate presence of a statistically significant difference across subjects. Significance levels: *$p<$ 0.05, **$p<$ 0.01, ***$p<$ 0.001.}
	\label{fig:barplots}
	\vspace{-0.1cm}
\end{figure*}

Figure~\ref{fig:barplots} presents the dimensionality reduction method results from Table~\ref{multiresults}, as well as a marked summary of these significance levels. We excluded CSP and FBCSP results from Figure~\ref{fig:barplots} since they were performed as baselines with no dimensionality reduction, and were usually statistically outperformed. For all analyses, we did not observe any significant differences by varying $d_y$ for MMI-LinT or MMI-NonLinT.


\section{Discussion}

We formulate a general definition for information theoretic feature transformation learning that we discuss to be Bayesian optimal for classification, and not based on feature selection heuristics. Derived by this definition, we present a linear and a nonlinear feature transformation framework. We evaluate the proposed approaches in decoding with respect to conventional CSP and FBCSP derived initial feature vectors as a baseline, statistical testing oriented feature ranking and selection methods ($R^2$ and SDA), as well as information theoretic feature ranking and selection methods (mRMR and MMI). For multi-class problems, we introduce a graphical model based hierarchical decoding framework, which can be considered as intuitively structured one-versus-rest classifiers. We believe that this hierarchical binary feature transformation learning approach is likely to expand conventional multi-class BCIs. Binary and multi-class decoding results on a four class motor imagery BCI task demonstrate statistically significant performance increases by feature transformation learning, with regards to state-of-the-art feature selection methods.

In discriminative model learning, feature selection is a sub-optimal approach towards the ultimate objective of maximizing mutual information by feature transformations. However, estimating this objective in Eq.~\ref{eq:objective} is challenging since it is simultaneously based on multiple continuous and discrete random variables. A related line of work tackles the problem of finding global solutions to a similar objective in mutual information based feature selection contexts \cite{Rodriguez:2010,Nguyen:2014}. There also exists some recent work on estimating mutual information for such discrete-continuous mixtures \cite{Ross:2014,Gao:2017}. One recent paper suggests measuring joint entropy among multiple variables in the reproducing kernel Hilbert space, thus enabling estimation of mutual information between discrete and continuous variables without explicit probability density function estimation \cite{Yu:2018}. Alternatively in this study, we propose a stochastic approximation to the problem, which was also previously studied with the same objective, using various non-parametric entropy estimation schemes \cite{Torkkola:2003,Chen:2008,Zhang:2010,Faivishevsky:2012}.

Proposed feature transformation learning approach can be interpreted as determining a manifold on which projections/transformations of the original extracted features carry maximal mutual information with their corresponding class labels, where this projection ideally provides an information theoretic upper bound with respect to any maximum mutual information based feature ranking and selection criteria. Consistently, any MMI feature selection algorithm can be seen as a constrained version of MMI-LinT with sparse orthonormal matrix linear projections. Hereby, we provide a broader definition which is likely to overcome potential shortcomings of feature selection. However, it is important to highlight the main drawback of the proposed method, that it does not maintain the direct neurophysiologically interpretable nature of feature ranking and selection. Feature transformations exploit synergies across initially constructed feature vectors, hence losing physical meanings. For instance in MMI-LinT, obtained features correspond to a combined measure of weighing across initial feature vectors. Nevertheless, this aligns with the hypothesis on the existence of large-scale cortical networks representative of specific tasks \cite{Mantini:2007,Bressler:2010}.

Stochastic mutual information gradients rely on estimating class conditional densities at each iteration. Here, a parametric (e.g., Gaussian) feature transformation choice would force the transformed data samples to follow a specified distribution, which may be restrictive when estimating mutual information \cite{Ang:2012,Lotte:2018}. Alternatively, kernel density estimations can performed over the two-dimensional transformed feature domain. Note that this approach is not equivalent to estimating high-dimensional raw EEG feature distributions with discretized kernels. Therefore these estimates in the transformed domain does not provide crude approximations over EEG features.

Commonly, BCI user intent inference pipelines contain subsequent pre-processing, feature extraction and selection steps. Instead of feature selection, the proposed method can simply replace this dimensionality reduction step as a stochastic MMI transformation estimator module. At training time, batch-wise iterative computations involve class conditional kernel density estimations, calculation of the gradient of Eq.~\ref{eq:smig}, and parameter updates for a specified number of epochs. Computational complexity increases linearly with the number of training data samples $n$ for a specific number of classes. At test time, computations simply include applying the transformation function (e.g., a single matrix multiplication in MMI-LinT).

A natural multi-class extension for CSP can be performed by combining pairwise CSP analyses for one-versus-rest classifiers as in our hierarchical approach, or directly generating features using multi-class labels (e.g., joint approximate diagonalization of class covariances) \cite{Dornhege:2004,Grosse-Wentrup:2008}. Our feature transformation learning formulation is also capable of directly learning with multi-class labels. However, we exploited a hierarchical decoding model for better level-wise binary feature learning, and reported our results in this framework for comparisons. Furthermore, the hierarchical graphical model based approach allows incorporating useful level transition priors for the BCI system \cite{Ozdenizci:2018}. Notably, our approach demonstrated a more significant advantage especially in the multi-class scenarios (i.e., Table~\ref{binary1resultstable} to Table~\ref{multiresults}). We believe this is an expected result given that one-versus-rest multi-class decoding can combine level-wise confounders. This is particularly interesting to observe the deficiencies and/or redundancies in features selections at pairwise comparisons which can accumulate.

The proposed approach did not reveal highly significant performance differences in some across sessions comparisons. There was also a drop in across sessions accuracies with respect to within session results, due to the challenging nature of the session-to-session transfer learning problem. This is an important observation to be emphasized regarding the practicality of the current approach, which can potentially be restricted by the amount of sessions (two) considered in our current experiments. We believe the current lack of generalizability can be a result of the across sessions unstability of EEG and the transformations we learn, which are based on single session-specific EEG data. One further exploration could be to exploit longitudinal BCI recordings performed over various sessions/days, and investigate the practicality of the our approach when multiple sessions' data are available for model training. Moreover, in such settings, one can explicitly impose session-invariance constraints to the feature transformation problem. This can be tackled in an adversarial learning framework which we are currently exploring based on our preliminary works \cite{Ozdenizci:2019b,Ozdenizci:2019}, where additional session-invariance constraints by an antagonistic objective regularizes feature learning pipelines. Another potential future direction is to consider information theoretic metric learning methods \cite{Davis:2007,Niu:2014}. This can be performed by learning distance metrics for transforms based on data covariance matrices (e.g., Mahalanobis distance) that utilizes a mutual information based cost.

Generalization and optimal exploitation of the information content in the extracted features with respect to their class labels is essential for discriminative model learning. We addressed the significance of this issue in the design of brain/neural interfaces. Given the significant evidence claiming that feature selection being potentially sub-optimal in model learning \cite{Guyon:2003,Erdogmus:2008,Torkkola:2008}, we argue that a feature transformation learning approach should be of important use in BCIs.

\section{Conclusion}

This work addresses the potential confounders caused by heuristic feature ranking and selection based dimensionality reduction methods that are widely used for brain interfaces. We extend this focus with a novel information theoretic feature transformation concept. We formulate a general definition for the feature learning problem, and present a linear and a nonlinear feature transformation approach derived by this definition. We further introduce a graphical model based, hierarchical binary feature transformation learning and decoding framework for multi-class scenarios. We empirically demonstrate that stochastic, mutual information based feature transformation learning significantly outperforms state-of-the-art feature selection heuristics, and yields significant insights for the growing field of neural interfaces.


\end{document}